\documentclass{article} % For LaTeX2e
\usepackage{nips13submit_e,times}
\usepackage{hyperref}
\usepackage{url}

\usepackage{array}
\usepackage{float}
\usepackage{multirow}
\usepackage[numbers]{natbib}
\usepackage{latexsym}
\usepackage{multicol}
\usepackage{eucal}
\usepackage{amsmath}
\usepackage{amsthm}
\usepackage{amsbsy}
\usepackage{amstext}
\usepackage{amssymb}
\usepackage{algpseudocode}
\usepackage{algorithm}
\usepackage{dsfont} %for mathds, blackboard nonbold
\usepackage{ifthen}
\usepackage{mathrsfs} %for math script
\usepackage{graphicx}
\usepackage{subcaption}
\usepackage{color}
\usepackage{todonotes}
\usepackage{microtype}
\usepackage{varwidth}

\newcommand{\cut}[1]{}

\DeclareMathOperator*{\argmax}{arg\,max}

\newcommand{\BigO}[1]{\ensuremath{\operatorname{O}\bigl(#1\bigr)}}

\title{Anytime Belief Propagation Using Sparse Domains}
\author{
Sameer Singh\\
University of Massachusetts\\
Amherst MA 01003\\
\texttt{sameer@cs.umass.edu} \\
\And
Sebastian Riedel\\
University College London\\
London UK\\
\texttt{s.riedel@cs.ucl.ac.uk} \\
\And
Andrew McCallum\\
University of Massachusetts\\
Amherst MA 01003\\
\texttt{mccallum@cs.umass.edu}
}

\nipsfinalcopy

\begin{document}

\maketitle
\cut{
\begin{abstract}
Belief Propagation has been widely used for marginal inference, however it is slow on problems with large-domain variables and high-order factors. Previous work provides useful approximations to facilitate inference on such models, but lacks important \emph{anytime} properties such as: 1) providing accurate and consistent marginals when stopped early, 2) improving the approximation when run longer, and 3) converging to the fixed point of BP. To this end, we propose a message passing algorithm that works on sparse (partially instantiated) domains, and converges to consistent marginals using dynamic message scheduling. The algorithm grows the sparse domains incrementally, selecting the next value to add using prioritization schemes based on the gradients of the marginal inference objective. Our experiments demonstrate local anytime consistency and fast convergence, providing significant speedups over BP to obtain low-error marginals: up to $25$ times on grid models, and up to $6$ times on a real-world natural language processing task.
\end{abstract}
}

% BP has been great, its been useful, even though its not converging to the right thing..
For marginal inference on graphical models, belief propagation (BP) has been the algorithm of choice due to impressive empirical results on many models. %based message-passing 
These models often contain many variables and factors, however the domain of each variable (the set of values that the variable can take) and the neighborhood of the factors is usually small.
When faced with models that contain variables with large domains and higher-order factors, BP is often intractable.
%
% why this happens, and what people have done about it
The primary reason BP is unsuitable for large domains is the cost of message computations and representation, which is in the order of the cross-product of the neighbors' domains.
Existing extensions to BP that address this concern \citep{coughlan07:dynamic,coughlan02:finding,ihler09:particle,noorshams11:stochastic,shen07:guided,sudderth03:nonparametric,yu07:efficient} use parameters that define the desired level of approximation, and return the approximate marginals at convergence. %the end of execution
% why none of them are anytime, and why anytime is really important
This results in poor \emph{anytime} behavior. % desired in the resource-constrained setting. 
Since these algorithms try to directly achieve the desired approximation, the marginals \emph{during} inference cannot be characterized, and are often inconsistent with each other.
Further, the relationship of the parameter that controls the approximation to the quality of the intermediate marginals is often unclear.
As a result, these approaches are not suitable for applications that require consistent, anytime marginals but are willing to trade-off error for speed, for example applications that involve real-time tracking or user interactions. 
There is a need for an anytime algorithm that can be interrupted to obtain consistent marginals corresponding to fixed points of a well-defined objective, and can improve the quality of the marginals over the execution period, eventually obtaining the BP marginals.

% what we propose as a class of algorithms
In this work we propose a novel class of message passing algorithms that compute accurate, anytime-consistent marginals. %that iterate between value instantiation and BP to
% two sentence summary
Initialized with a sparse domain for each variable, the approach alternates between two phases: (1) augmenting values to sparse variable domains, and (2) converging to a fixed point of the approximate marginal inference objective as defined by these sparse domains.
%We instantiate sparse, partial domains for each variable, but ensure that message passing converges to a fixed point on the approximate objective as defined over these restricted domains.
We tighten our approximate marginal inference objective by selecting the value to add to the sparse domains by estimating the impact of adding the value to the variational objective; this is an accurate prioritization scheme that depends on the instantiated domains and requires runtime computation. % for out of sparse domain values.
We also provide an alternate prioritization scheme based on the gradient of the primal objective that can be computed a priori, and provides constant time selection of the value to add. % to the To tighten the approximation, values are added to the instantiated domains, followed by message passing over these updated domains. 
To converge to a fixed point of the approximate marginal objective, we perform message passing on the sparse domains. % to obtain consistent marginals
Since naive schedules that update messages in a round robin or random fashion are wasteful, we use residual-based dynamic prioritization~\citep{elidan06:residual}. %iterate through factors, zation has been used
%Although these approaches are inefficient on large domains, they are suitable for the small sizes of our instantiated domains.
%existing domains.
Inference can be interrupted to obtain consistent marginals at a fixed point defined over the instantiated domains, and longer execution results in more accurate marginals, eventually optimizing the BP objective.

%-----------------------------------------------------------
%-----------------------------------------------------------
%-------------------      MARGINAL INF     -----------------
%-----------------------------------------------------------
%-----------------------------------------------------------
\section{Marginal Inference for Undirected Graphical Models}
\label{sec:prelims}

\global\long\def\e{e}

\global\long\def\s{\mathbf{x}}

\global\long\def\parse{y}

\global\long\def\trees{\mathcal{Y}}

\global\long\def\z{\mathbf{z}}

\global\long\def\w{\boldsymbol{\theta}}

\global\long\def\f{\mathbf{f}}

\global\long\def\rc{r}

\global\long\def\mod{\text{mod}}

\global\long\def\gp{\text{gp}}

\global\long\def\arc{}

\global\long\def\cand{P}

\global\long\def\temp{t}

\global\long\def\Templates{\mathcal{T}}

\global\long\def\inc{\delta^{-}}

\global\long\def\outg{\delta^{+}}

\global\long\def\flowdual{\text{flow}}

\global\long\def\defeq{\triangleq}

\global\long\def\indi{\mathbb{I}}

\global\long\def\Z{\mathcal{Z}}

\global\long\def\M{\mathcal{M}}

\global\long\def\L{\mathcal{L}}

\global\long\def\b{\mathbf{b}}

\global\long\def\flows{\boldsymbol{\phi}}

\global\long\def\head{\text{h}}

\global\long\def\Vo{V_{-0}}

\global\long\def\duals{\boldsymbol{\lambda}}

\global\long\def\u{\mathbf{u}}

\global\long\def\classof{\text{cl}}

\global\long\def\sgn{\text{sgn}}

\global\long\def\A{\mathbf{A}}

%\global\long\def\marg{\boldsymbol{\tau}}
\global\long\def\msg{\boldsymbol{m}}

\global\long\def\x{\mathbf{x}}
\global\long\def\v{\mathbf{v}}

\global\long\def\X{\mathcal{X}}
\global\long\def\S{\mathcal{S}}
\global\long\def\D{\mathcal{D}}
\global\long\def\N{\mathcal{N}}
\global\long\def\F{\mathcal{F}}

\global\long\def\Hb{H_{B}}

\global\long\def\f{\boldsymbol{\phi}}

\global\long\def\E{\text{E}}

\global\long\def\marg{\boldsymbol{\mu}}

Let $\x$ be a random vector where each $x_{i}\in\x$ takes a value $v_i$ from  domain $\D$. %\defeq\left\{ 0,1,\ldots,D-1\right\} 
An assignment to a subset of variables $\x_c\subseteq\x$ is represented by $\v_c\in\D^{|\x_c|}$.
A factor graph~\citep{kschischang01:factor} is defined by a bipartite graph over the variables $\x$ and a set of factors $f\in\F$ (with neighborhood $\x_f\equiv\N(f)$).
Each factor $f$ defines a scalar function $\f_f$ over the assignments $\v_f$ of its neighbors $\x_f$, defining the distribution: \\%as induced by $G$:
$\displaystyle 
p(\v)\defeq\frac{1}{Z}\exp\sum_{f\in\F}\f_f(\v_f)\text{, where } Z\defeq\sum_{\v'\in\D^{n}}\exp\sum_{f\in\F}\f_f(\v'_f)
$. %\nonumber
Inference is used to compute the variable marginals $p(v_i) = \sum_{\v/x_i} p(\v)$ and the factor marginals $p(\v_f) = \sum_{\v/\x_f} p(\v)$.
When performing approximate variational inference, we represent the approximate marginals $\marg\equiv(\marg_\X,\marg_\F)$ that contain elements for every assignment to the variables $\marg_\X\equiv\mu_i(v_i), \forall x_i,v_i\in \D$ and factors $\marg_\F\equiv\mu_f(\v_f), \forall f,\v_f\in\D^{|\x_f|}$.
Minimizing the KL divergence between the desired and approximate marginals results in: % optimization:
%\begin{equation}
%\[
{\small\quad$\displaystyle
\max_{\marg\in\M}\sum_{f\in\F}\sum_{\v_f}\mu_f(\v_f)\f_f(\v_f)+H\left(\marg\right)
$}, %\label{eq:variational}
%\]
%\end{equation}
where $\M$ is the set of \emph{realizable} mean vectors $\marg$, and $H\left(\marg\right)$ is the entropy of the distribution that yields $\marg$.
%The maximizer $\marg^{*}\in\M$ is the mean vector that corresponds to the result of marginal inference. %for Equation~\eqref{eq:variational}
%
%\textbf{Belief Propagation: }
Both the polytope $\M$ and the entropy $H$ need to be approximated in order to efficiently solve the maximization.

{\bf Belief propagation (BP)} approximates $\M$ using the \emph{local polytope}:\\
%\begin{eqnarray}
{\small$
\L\defeq\biggl\{ \marg\geq0, ~~~\forall f\in \F:\sum_{\v_f}\mu_f\left(\v_f\right)=1,
\forall f,i\in \N\left(f\right),v_{s}:\sum_{\v'_{f},v_{i}'=v_s}\mu_{f}\left(\v'_{f}\right)=\mu_{i}\left(v_s\right)\biggr\} %\nonumber
$}
%\end{eqnarray}
and entropy using Bethe approximation:
%\begin{eqnarray}
%\[
$\Hb\left(\marg\right)\defeq\sum_{f}H\left(\marg_{f}\right)-\sum_{i}\left(d_{i}-1\right)H\left(\marg_{i}\right)$,
%\]
%\end{eqnarray}
leading to: % the following optimization problem:
{\small\begin{equation}
\max_{\marg\in\L}\sum_{f\in\F}\sum_{\v_f}\mu_f(\v_f)\f_f(\v_f)+\Hb\left(\marg\right) \label{eq:bp:primal}
\end{equation}}
The Lagrangian relaxation of this optimization is:
{\small\begin{equation}
%\[
L_\text{BP}\left(\marg,\duals\right)\defeq\sum_{f\in\F}\sum_{\v_f}\mu_f(\v_f)\f_f(\v_f) +\Hb\left(\marg\right) %\nonumber\\ 
+\sum_{f}\lambda_fC_f\left(\marg\right) %\sum_{v_i} &+&
+\sum_{f}\sum_{i\in N\left(f\right)}\sum_{v_i}\lambda_{fi}^{v_i}C_{f,i,v_i}\left(\marg\right) \label{eq:bp:dual}
%\]
\end{equation}}
where $C_{f,i,v_i}=\mu_i(v_i)-\sum_{\v_f/x_i}\mu_f(\v_f)$ and $C_f=1-\sum_{\v_f}\mu_f(\v_f)$ are the constraints that correspond to the local polytope $\L$.
BP messages correspond to the dual variables, i.e. $\msg_{fi}(v_i)\propto\exp\lambda_{fi}^{v_i}$.
If the messages converge, \citet{yedidia00:generalized} show that the marginals correspond to a $\marg^{*}$ and $\duals^{*}$ at a saddle point of $L_{\text{BP}}$, i.e. $\nabla_{\marg}L_{\text{BP}}\left(\marg^{*},\duals^{*}\right)=0$ and $\nabla_{\duals}L_{\text{BP}}\left(\marg^{*},\duals^{*}\right)=0$. 
In other words: at convergence BP marginals are locally consistent and locally optimal. % set of marginals. 
BP is not guaranteed to converge, or to find the global optimum if it does, however it often converges and produces accurate marginals in practice~\citep{murphy99:loopy}.

%-----------------------------------------------------------
%-----------------------------------------------------------
%-------------------      ANYTIME    BP      ---------------
%-----------------------------------------------------------
%-----------------------------------------------------------
\section{Anytime Belief Propagation} % for Large Domains
\label{sec:proposed}

%Why variables with large domains and higher-order factors are really important.
Graphical models are often defined over variables with large domains and factors that neighbor many variables.
%Why does BP face trouble with variables with large domains.
Message passing algorithms perform poorly for such models since the complexity of message computation for a factor is $\BigO{|\D|^{|\N_f|}}$ where $\D$ is the domain of the variables. %, and $|\N_f|$ is the neighborhood size. % of the factors.
Further, if inference is interrupted, the resulting marginals are not locally consistent, nor do they correspond to any fixed point of a well-defined objective. %n interpretable and
%how a number of existing approaches that rely on fast message computation get slower.
Here, we describe an algorithm that meets the following desiderata:
(1)~anytime property that results in consistent marginals, % and accurate marginals (over partially instantiated domains),
(2)~more iterations improve the accuracy of marginals, 
and (3) convergence to BP marginals (as obtained at a fixed point of BP).
%To obtain accurate marginals quickly, we utilize the current state of inference to instantiate high-likelihood values before picking low probability ones.

%\inline{point to subsection ``modules''}
%Marginal inference is often cast as optimizing approximations of the free energy over a set of locally consistent marginals. 
%BP performs this optimization through means of message passing, and while doing so it is computing marginals that are both locally consistent and favorable by the free energy approximation. 
Instead of directly performing inference on the complete model, our approach maintains \emph{partial} domains for each variable.
Message passing on these sparse domains converges to a fixed point of a well-defined objective (Section~\ref{sec:proposed:bp}).
This is followed by incrementally \emph{growing} the domains (Section~\ref{sec:proposed:grow}), and resuming message passing on the new set of domains till convergence.
At any point, the marginals are close to a fixed point of the sparse BP objective, and we tighten this objective over time by growing the domains.
If the algorithm is not interrupted, entire domains are instantiated, and the marginals converge to a fixed point of the complete BP objective. % (Equation~\ref{eq:bp:dual}). allowed to execute without interruption

\subsection{Belief Propagation with Sparse Domains}
\label{sec:proposed:bp}

First we study the propagation of messages when the domains of each variables have been partially instantiated (and are assumed to be fixed here).
Let $\S_i\subseteq D, |\S_i|\geq1$ be the set of values associated with the instantiated domain for variable $x_i$.
During message passing, we \emph{fix} the marginals corresponding to the non-instantiated domain to be zero, i.e. $\forall v_i\in \D-\S_i, \mu_i(v_i)=0$.
By setting these values in the BP dual objective \eqref{eq:bp:dual}, we obtain the optimization defined only over the sparse domains:
% message passing with these domains corresponds to the optimization defined only over the sparse domains:
%, and is represented by the following approximation:
{\small\begin{eqnarray}
L_\text{SBP}\left(\marg,\duals,\S\right)\defeq\sum_{f}\sum_{\v_f\in\S_f}\mu_f(\v_f)\f_f(\v_f) +\Hb\left(\marg\right) %\nonumber\\
+\sum_{f}\lambda_fC_f\left(\marg\right)%\nonumber\\ \sum_{v_i\in\S_i}
+\sum_{f}\sum_{i\in N\left(f\right)}\sum_{v_i\in\S_i}\lambda_{fi}^{v_i}C_{f,i,v_i}\left(\marg\right)\label{eq:sbp}
\end{eqnarray}}
Note that $L_{\text{SBP}}(\marg,\duals, \D^n)=L_{\text{BP}}(\marg,\duals)$.
Message computations for this approximate objective, including the summations in the updates, are defined sparsely over the instantiated domains.
In general, for a factor $f$, the computation of its outgoing messages requires $\BigO{\prod_{x_i\in\N_f}|\S_i|}$ operations, as opposed to $\BigO{|D|^{|\N_f|}}$ for whole domains. %, leading to faster inference for the approximation.
Variables for which $|\S_i|=1$ are treated as \emph{observed}. %deterministic messages on the sparse values.
%\inline{show updates with sparse domains?}

\subsection{Growing the Domains}
\label{sec:proposed:grow}

As expected, BP on sparse domains is much faster than on whole domains, however it is optimizing a different, approximate objective.
The approximation can be tightened by growing the instantiated domains, that is, as the sparsity constraints of $\mu_i(v_i)=0$ are removed, we obtain more accurate marginals when message passing for newly instantiated domain converges.
Identifying \emph{which} values to add is crucial for good anytime performance, and we propose two approaches here based on the gradient of the variational and the primal objectives. %in this section. % to the instantiated domains

\textbf{Dynamic Value Prioritization: }
%\label{sec:proposed:dynamic}
%
When inference with sparse domains converges, we obtain marginals that are locally consistent, and define a saddle point of Eq~\eqref{eq:sbp}. % to Belief Propagation on the whole domain (Equation~\ref{eq:bp:dual}).
We would like to add the value $v_i$ to $\S_i$ for which removing the constraint $\mu_i(v_i)=0$ will have the most impact on the approximate objective $L_{\text{SBP}}$.
In other words, we select $v_i$ for which the gradient $\frac{\partial L_{\text{SBP}}}{\partial \mu_i(v_i)}|_{\mu_i(v_i)=0}$ is largest. % (tighten as much as possible).
%This can be computed by computing the gradient of the free energy when a particular value $v_i$ is added to the current domain $\S_i$ of variable $x_i$.
From \eqref{eq:sbp} we derive $\frac{\partial L_{SBP}}{\partial\mu_i(v_i)} = (d_i-1)(1+\log\mu_i(v_i))+\sum_{f\in\N(x_i)}\lambda_{fi}^{v_i}$.
Although $\log\mu_i(v_i)\rightarrow-\infty$ when $\mu_i(v_i)\rightarrow0$, we ignore the term as it appears for all $i$ and $v_i$\footnote{Alternatively,  approximation to $L_{\text{SBP}}$ that replaces the variable entropy $-\sum_pp\log p$ with its second order Taylor approximation $\sum_pp(1-p)$. The gradient at $\mu_i(v_i)=0$ of the approximation is $d_i+\sum_{f\in\N(x_i)}\lambda_{fi}^{v_i}$.}. %, i.e. same as in \eqref{eq:sbp:dyn}.}.
%We also ignore $d_i$ for models where the variable neighborhood sizes are similar.
The rest of the gradient is the priority:
{\small$\pi_i(v_i) = d_i+\displaystyle\sum_{f\in\N(x_i)}\lambda_{fi}^{v_i}$}. %\label{eq:sbp:dyn}
%\pi_i(v_i) &=& \sum_{f\in\N(x_i)}\lambda_{fi}^{v_i}\label{eq:sbp:dyn}
%\end{eqnarray}
%
Since $\lambda_{fi}^{v_i}$ is undefined for $v_i\notin\S_i$, we estimate it by performing a single round of message update over the sparse domains.
%\begin{eqnarray}
%$\lambda_{fi}^{v_i}= \log\sum_{\v_f\in\S_f/x_i} \exp\left\{\f_f(\v_f) + \sum_{x_j\in\N_f/x_i}\lambda_{jf}^{\v_f(j)}\right\}$. %\nonumber
%\end{eqnarray}
To compute priority of all values for a variable $x_i$, this computation requires %$\BigO{\sum_f|\D-\S_i|\prod_{\N_f/x_i}|\S_i|}$, i.e. 
an efficient $\BigO{|\D||\S|^{\N_f-1}}$.
Since we need to identify the value with the highest priority, we can improve this search by sorting factor scores $\f$, and %, though the worst case complexity remains the same.
further, we only update the priorities for the variables that have participated in message passing.

\textbf{Precomputed Priorities of Values: }
%\label{sec:proposed:fixed}
%
%\inline{mention this is an alternative}
Although the dynamic strategy selects the value that improves the approximation the most, it also spends time on computations that may not result in a corresponding benefit.
As an alternative, we propose a prioritization that precomputes the order of the values to add; even though this does not take the current beliefs into account, the resulting savings in speed may compensate.
Intuitively, we want to add values to the domain that have the highest marginals in the final solution.
Although the final marginals cannot be computed directly, we estimate them by enforcing a single constraint $\mu_{i}(v_i)=\sum_{\v_{f}/x_i}\mu_{f}(\v_{f})$ and performing greedy coordinate ascent for each $f$ on the primal objective in \eqref{eq:bp:primal}. We set the gradient w.r.t. $\mu_f(\v_f)$ to zero to obtain:
%\begin{equation}
{\small\quad$\displaystyle
\pi_i(v_i) = \hat\mu_i(v_i)=\sum_{\v'_{f},v_{i}'=v_s}\hat\mu_{f}\left(\v'_{f}\right)%\nonumber\\
=\sum_{f\in\N(x_i)}\log \sum_{\v_f\in\D_f\x_i}\exp \f_f(\v_f)
$}. %\label{eq:sbp:fixed}
%\end{equation}
%\vskip -5mm
This priority can be precomputed and identifies the next value to add in constant time. % to 

\subsection{Dynamic Message Scheduling}
\label{sec:proposed:rbp}

%\inline{say this is orthogonal to growing domains}
After the selected value has been added to its respective domain, we perform message passing as described in Section~\ref{sec:proposed:bp} to converge to a fixed point of the new objective. %compensate for this relaxation. %locally consistent marginals and converge.
To focus message updates in the areas affected by the modified domains, we use dynamic prioritization amongst messages~\citep{elidan06:residual,sutton07:improved} with the dynamic range of the change in the messages (\emph{residual}) as the choice of the message norm~\citep{ihler05:loopy}. %, as explored by \citet{ihler05:loopy} and \citet{sutton07:improved}.
Formally:
%\[ %begin{equation}
%\pi(f) = \max_{x_i\in\N_f}||\msg_{fi}-\msg'_{fi}||_{\infty}
$\displaystyle\pi(f) = \max_{x_i\in\N_f}\max_{v_i,v_j\in S_i}\log\cfrac{e(v_i)}{e(v_j)}, ~~~e=\cfrac{\msg_{fi}}{\msg'_{fi}}$.
%\] %end{equation}
As shown by \citet{elidan06:residual}, residuals of this form bound the reduction in distance between the factor's messages and their fixed point, allowing their use in two ways: first, we pick the highest priority message since it indicates the part of the graph that is least locally consistent. 
Second, the maximum priority is an indication of convergence and consistency; a low max-residual implies a low bound on the distance to convergence. 

%-----------------------------------------------------------
%-----------------------------------------------------------
%-------------------       EXPERIMENTS       ---------------
%-----------------------------------------------------------
%-----------------------------------------------------------
\section{Experiments}
\label{sec:experiments}

Our primary baseline is \emph{Belief Propagation} (\textbf{BP}) using random scheduling.
We also evaluate \emph{Residual BP} (\textbf{RBP}) that uses dynamic message scheduling. %, as described in Section~\ref{sec:proposed:rbp}. %on the change in the outgoing messages 
Our first baseline that uses sparsity, \emph{Truncated Domain} (\textbf{TruncBP}), is initialized with domains that contain a fixed fraction of values ($0.25$) selected according to precomputed priorities (Section~\ref{sec:proposed:grow})
and are not modified during inference. %, and perform message passing as in Section~\ref{sec:proposed:bp}. 
We evaluate three variations of our framework.
\emph{Random Instantiation} (\textbf{Random}) is the baseline that the value to be added at random, followed by priority based message passing.
Our approach that estimates the gradient of the dual objective is \emph{\bf Dynamic}, while the approach that precomputes priorities is \emph{\bf Fixed}. %,  and is the version of our algorithm that utilizes the incoming messages for value priorities, as described in.

\begin{figure}
\centering
\begin{minipage}{.45\textwidth}
  \centering
  \includegraphics[width=0.8\linewidth]{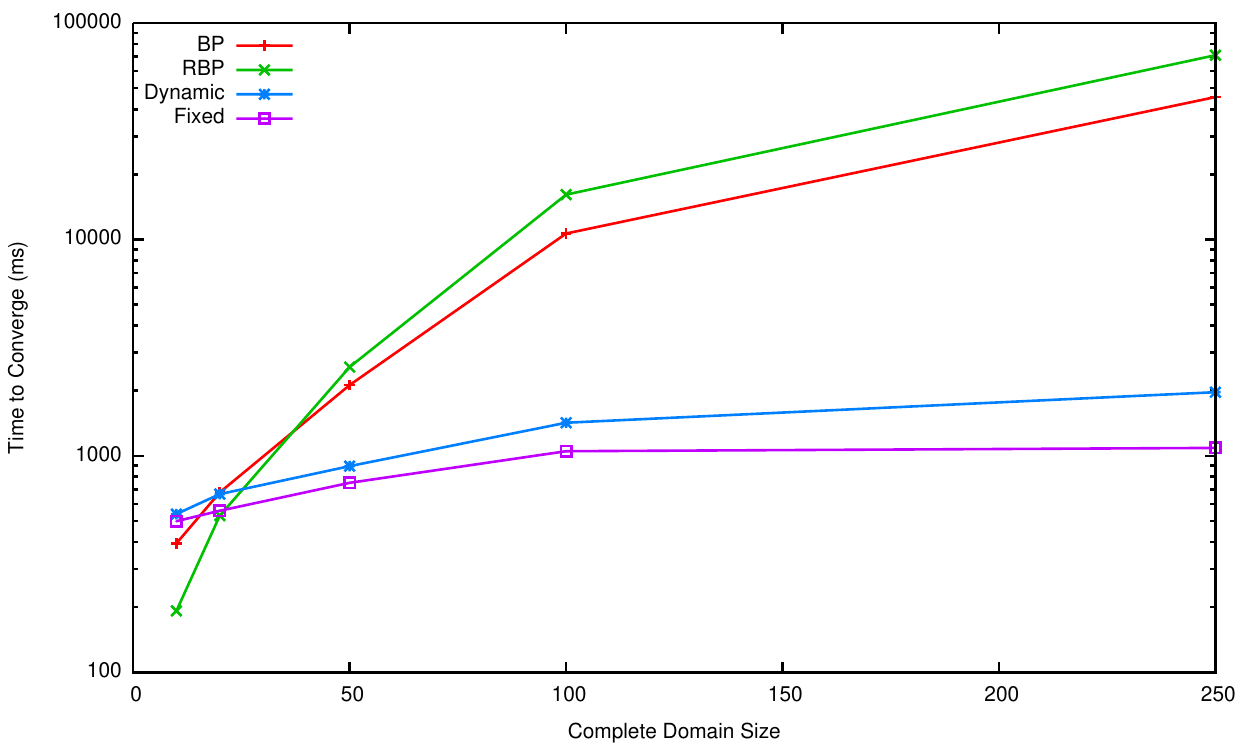}
  \captionof{figure}{\small{\bf Convergence time for different domains:} to $L_2<10^{-4}$ over $10$ runs of $5\times5$ grids.}
  \label{fig:grid:scalability}
\end{minipage}%
\qquad
\begin{minipage}{.4\textwidth}
                \centering
                \raisebox{16mm}{\small
			    \begin{tabular}{|l|c|c|c|}
			      \hline
			      \bf \# Entities & 4 & 6 & 8 \\
			      \bf \# Vars & 16 & 36 & 64 \\
			      \bf \# Factors & 28 & 66 & 120 \\
			      \hline
			      \bf BP & 41,193 & 91,396 & 198,374\\
			      \bf RBP & 54,577 & 117,850 & 241,870\\
			      \bf Fixed & 24,099 & 26,981 & 49,227\\
			      \bf Dynamic & 24,931 & 36,432 & 41,371 \\
			      \hline
			    \end{tabular}
			    }
  		  \captionof{figure}{\small\label{tab:jnt:times} {\bf Joint Information Extraction:} Avg time taken (ms) to $L_2<0.001$}
\end{minipage}
\vskip-3mm
\end{figure}

\textbf{Grids: }
%Our first set of experiments evaluate on synthetic grid models.
Our first testbed for evaluation consists of $5\times 5$ and $10\times10$ grid models (with domain size of $L=10,20,50,100,250$), consisting of synthetically generated unary and pairwise factors. %consisting variables 
\begin{figure*}[tb]
\centering
\begin{subfigure}[b]{0.32\textwidth}
                \centering
                \includegraphics[width=\textwidth]{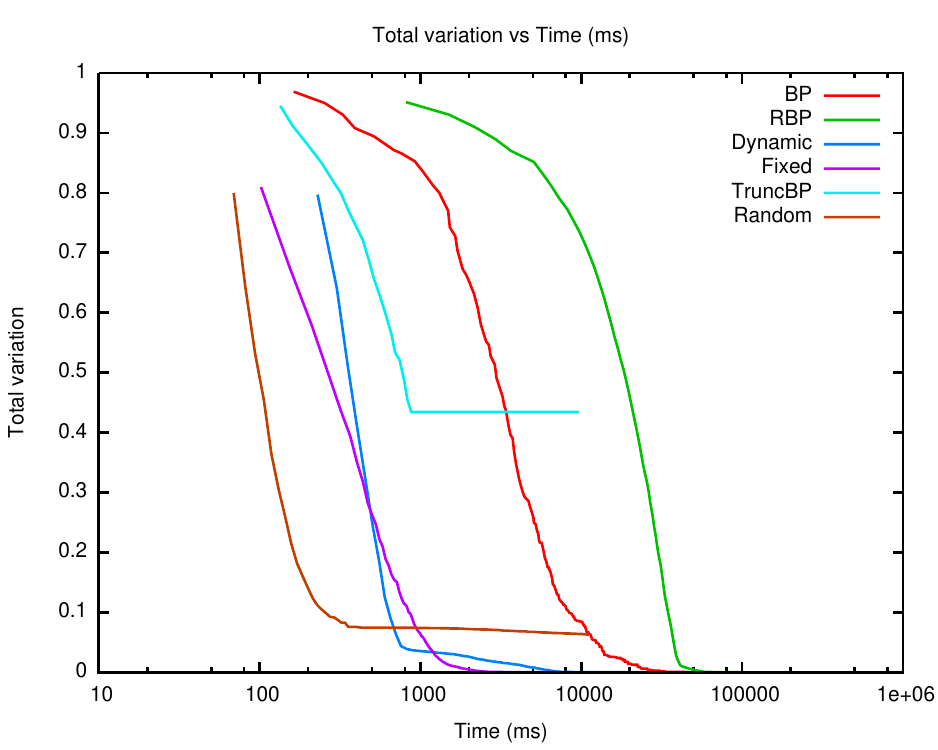}
                \caption{Total Variation Distance}
                \label{fig:grid:metrics:l1}
        \end{subfigure}
\begin{subfigure}[b]{0.32\textwidth}
                \centering
                \includegraphics[width=\textwidth]{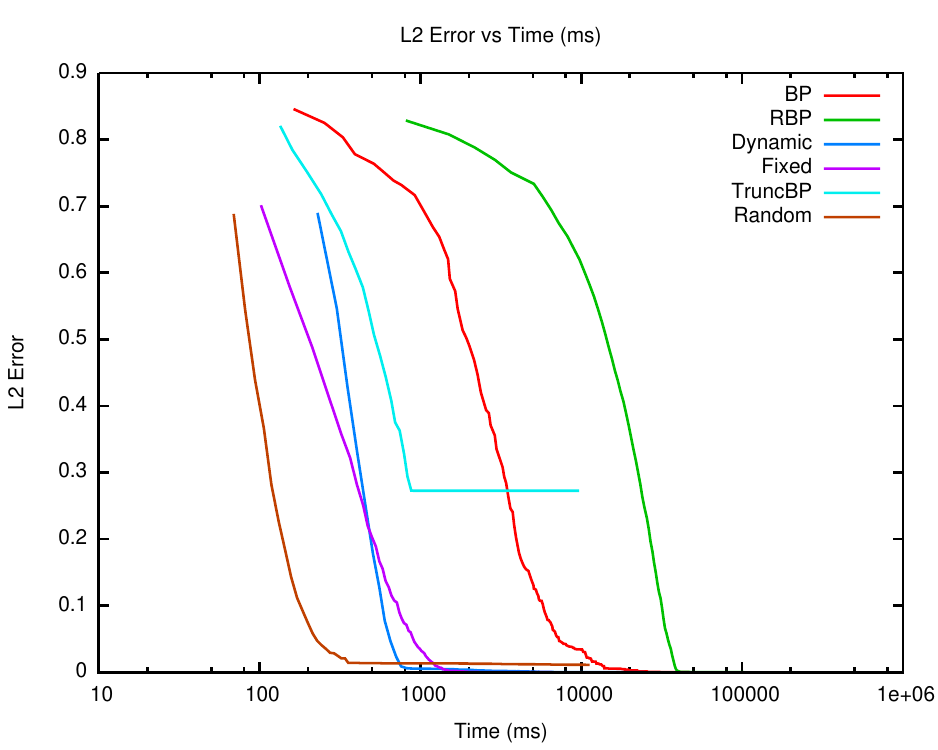}
                \caption{L2 Error}
                \label{fig:grid:metrics:l2}
        \end{subfigure}
\begin{subfigure}[b]{0.32\textwidth}
                \centering
                \includegraphics[width=\textwidth]{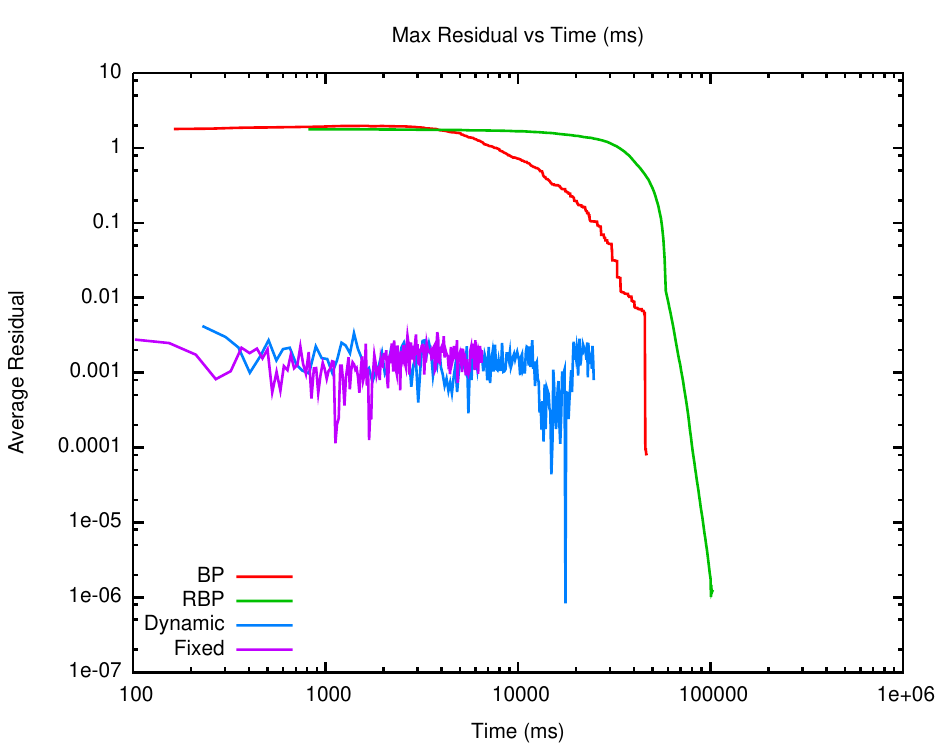}
                \caption{Average Residual in Messages}
                \label{fig:grid:res}
        \end{subfigure}
\caption{\small{\bf Runtime Analysis:} for 10$\times$10 grid with domain size of $100$, averaged over $10$ runs. }
\label{fig:grid:metrics}
\vskip-3mm
\end{figure*}
%
%We compare a number of error metrics in Figure~\ref{fig:grid:metrics}. % and \ref{fig:nlp:metrics}. as a function of running time
The runtime error for our approaches compared against the marginals obtained by BP at convergence (Figure~\ref{fig:grid:metrics}) is significantly better than BP; up to $\bf 12$ times faster to obtain $L_2$ error of $10^{-7}$. % for the grids. %, and up to $X$ times faster to for joint inference. % BP = 45536,45821, RBP = 66954, Fixed = 3737, Dynamic = 10008
TruncBP is efficient, however converges to an inaccurate solution, suggesting that prefixed sparsity in domains is not desirable.
Similarly, Random is initially fast, since adding \emph{any} value has a significant impact, however as the selections become crucial, the rate of convergence slows down considerably. % in reducing the error
Although both Fixed and Dynamic provide desirable trajectories, Fixed is much faster initially due to constant time growth of domains. However as messages and marginals become accurate, the dynamic prioritization that utilizes them eventually overtakes the Fixed approach.  
%\inline{compare fixed and dynamic}
%These experiments also demonstrate the inefficiency of RBP for models with large domains. %utility of regular BP over
%
%\textbf{Runtime Behavior: }
To examine the anytime local consistency, we examine the average residuals in Figure~\ref{fig:grid:res} % of our anytime predictions
%Since \emph{max}-residuals are an upper-bound on the reduction in distance to the locally-consistent BP fixed point, 
since low residuals imply a consistent set of marginals for the objective defined over the instantiated domain.
%Since we converge to a consistent solution on partially instantiated domains before growing domains, 
Our approaches demonstrate low residuals throughout, % the runtime.
%Specifically, the spikes in the figure occur at points of domain growth, and by passing messages in the neighborhood, the local messages are made consistent, reducing the residuals.
while the residuals for existing techniques remain significantly higher (note the log-scale), lowering only near convergence.
%\textbf{Varying the Domain Size: }
When the total domain size is varied in Figure~\ref{fig:grid:scalability}, % in Figure~\ref{fig:grid:scalability}.
%We run $10$ runs of $5\times5$ synthetic grids and observe the time to converge for different domain sizes to a low $L_2$ error in Figure~\ref{fig:grid:scalability}.
we observe that although our proposed approaches are slower on problems with small domains, they obtain significantly higher speedups on larger domains ($\bf 25-40$ times on $250$ labels).

\textbf{Joint Information Extraction: }
%\label{sec:joint}
%
We also evaluate on the real-world task of joint entity type prediction and relation extraction for the entities that appear in a sentence. 
The domain sizes for entity types and relations is $42$ and $24$ respectively, resulting in $42,336$ neighbor assignments for joint factors (details omitted due to space).
Figure~\ref{tab:jnt:times} shows the convergence time averaged over $5$ runs.
%as compared to the marginals at convergence.
For smaller sentences, sparsity does not help much since BP converges in a few iterations. %relatively 
However, for longer sentences containing many more entities, we observe a significant speedup (up to $\bf 6$ times). % on the largest models

\section{Conclusions} % and Future Work}
\label{sec:conclusions}

% Summarize Contributions
In this paper, we describe a novel family of \emph{anytime} message passing algorithms designed for marginal inference on problems with large domains.
%Instead of instantiating complete domains, 
The approaches maintain sparse domains, and efficiently compute updates that quickly reach the fixed point of an approximate objective by using dynamic message scheduling. %defined on the sparse domains
%This results in marginals that are close to the primal feasible region for the duration of inference.
Further, by growing domains based on the gradient of the objective, we improve the approximation iteratively, eventually obtaining the BP marginals.
%Results on synthetic and real world models demonstrate speedups of up to $25$ times over BP.

%\clearpage
%{\small
\bibliographystyle{plainnat}
\bibliography{../../../bibtex/sameer}  % sigproc.bib is the name of the Bibliography in this case
%}

\clearpage
\appendix
\section{Algorithm}
\label{sec:proposed:algo}
The proposed approach is outlined in Algorithm~\ref{alg:anytime}.
%As an input, the algorithm accepts a factor graph $G$.
We initialize the sparse domains using a single value for each variable with the highest priority. 
The domain priority queue ($Q_d$) contains the priorities for the rest of the values of the variables, which remain fixed or are updated depending on the prioritization scheme of choice (Section~\ref{sec:proposed:grow}).
%, and an empty message priority queue $Q_m$.
Message passing uses dynamic message prioritization maintained in the message queue $Q_m$.
Once message passing has converged to obtain locally-consistent marginals (according to some small $\epsilon$), we select another value to add to the domains using one of the value priority schemes, and continue till all the domains are fully-instantiated.
If the algorithm is interrupted at any point, we return either the current marginals, or the last converged, locally-consistent marginals. % and optimize the objective as defined over the instantiated domains.  that are locally consistent
We use a heap-based priority queue for both messages and domain values, in which update and deletion take $O(\log n)$, where $n$ is often smaller than the number of factors and total possible values. %removing

\begin{algorithm}[hb]
	\centering
	\caption{Anytime Belief Propagation\label{alg:anytime}}
	\begin{algorithmic}[1]
					%\Begin
					\State $\forall x_i, \S_i \gets \{v_i\}$ \Comment{where $v_i = \displaystyle\argmax_{v_s}\pi_i(v_s)$}
					%\State where $v_i = \displaystyle\argmax_{v_s}\pi_i(v_s)$
					\State $Q_d \oplus \left \langle (i, v_i),\pi_i(v_i)\right \rangle$ \Comment{$\forall x_i, v_i \in\D_i-\S_i$}
					\State $Q_m = \{\}$
					\While{$|Q_d| > 0$} \Comment{Domains are still partial}
					    \State \Call{GrowDomain}{$\S$, $Q_d$} \Comment{Add a value to a domain, Algorithm~\ref{alg:grow}}
					    \State \Call{ConvergeUsingBP}{$\S$, $Q_m$}\Comment{Converge to a fixed point, Algorithm~\ref{alg:sparse}}
					\EndWhile \Comment{Converged on full domains}
					%\End
	\end{algorithmic}
\end{algorithm}
\begin{algorithm}[hb]
	\centering
	\caption{Growing by a single value (Section~\ref{sec:proposed:grow})\label{alg:grow}}
	\begin{algorithmic}[1]
					\Procedure{GrowDomain}{$\S, Q_d$}
					    \State $(i, v_p)\leftarrow Q_d.pop$ \Comment{Select value to add}
					    \State $\S_i \leftarrow \S_i \cup \{v_p\}$ \Comment{Add value to domain}
					    \For{$f\leftarrow \mathcal{N}(x_i)$}
					        \State $Q_m \oplus \langle f,\pi(f) \rangle$ \Comment{Update msg priority}
					    \EndFor
					\EndProcedure
	\end{algorithmic}%
\end{algorithm}
\begin{algorithm}[hb]
	\centering
	\caption{BP on Sparse Domains (Sect.~\ref{sec:proposed:bp}, \ref{sec:proposed:rbp})\label{alg:sparse}}
	\begin{algorithmic}[1]
					\Procedure{ConvergeUsingBP}{$\S, Q_m$}
					    \While{$max(Q_m) > \epsilon$}
					        \State $f\leftarrow Q_m.pop$ \Comment{Factor with max priority}
					        \State Pass messages from $f$
					        \For{$x_j \leftarrow \N_f; f' \leftarrow \N(x_j)$}
					            \State $Q_m \oplus\langle f',\pi(f') \rangle$  \Comment{Update msg priorities}
					            \State $Q_d \oplus\langle(k,v_q),\pi_k(v_k)\rangle$\Comment{$\forall x_k\in\N_{f'},\forall v_k$}
					        \EndFor
					    \EndWhile \Comment{Converged on sparse domains}
					\EndProcedure
	\end{algorithmic}
\end{algorithm}

\end{document}